\documentclass[runningheads]{src/llncs}

\usepackage[mobile]{src/eccv}

\usepackage{graphicx}
\usepackage{booktabs}
\usepackage{multirow}
\usepackage{epigraph}
\usepackage[accsupp]{axessibility}  %

\usepackage[pagebackref,breaklinks,colorlinks,citecolor=eccvblue]{hyperref}
\usepackage{orcidlink}

\newcommand{\bA}{\mathbf{A}}

\newcommand{\bc}{\mathbf{c}}

\newcommand{\bI}{\mathbf{I}}

\newcommand{\bL}{\mathbf{L}}
\newcommand{\bM}{\mathbf{M}}

\newcommand{\bv}{\mathbf{v}}

\newcommand{\nN}{\mathbb{N}}

\newcommand{\nR}{\mathbb{R}}

\newcommand{\cE}{\mathcal{E}}

\newcommand{\cG}{\mathcal{G}}

\newcommand{\cL}{\mathcal{L}}
\newcommand{\cM}{\mathcal{M}}

\newcommand{\cO}{\mathcal{O}}
\newcommand{\cP}{\mathcal{P}}

\newcommand{\cR}{\mathcal{R}}
\newcommand{\cS}{\mathcal{S}}

\newcommand{\cV}{\mathcal{V}}

\newcommand{\figref}[1]{Fig.~\ref{#1}}
\newcommand{\secref}[1]{Section~\ref{#1}}

\newcommand{\tabref}[1]{Table~\ref{#1}}

\makeatletter
\DeclareRobustCommand\onedot{\futurelet\@let@token\@onedot}
\def\@onedot{\ifx\@let@token.\else.\null\fi\xspace}
\def\eg{e.g\onedot} 
\def\ie{i.e\onedot}

\makeatother

\newcommand{\boldparagraph}[1]{\vspace{0.2cm}\noindent{\bf #1.}}

\definecolor{darkgreen}{rgb}{0,0.7,0}
\definecolor{darkyellow}{rgb}{0.8,0.8,0}
\definecolor{bittersweet}{rgb}{1.0, 0.44, 0.37}
\definecolor{amber}{rgb}{1.0, 0.49, 0.0}
\definecolor{lgray}{rgb}{0.83,0.83,0.83}

\definecolor{color_unlabled}{rgb}{0.0,0.0,0.0}
\definecolor{color_vehicle}{rgb}{0.0,0.0,0.56}
\definecolor{color_road}{rgb}{0.5,0.25,0.5}
\definecolor{color_redlight}{rgb}{1.0,0.0,0.0}
\definecolor{color_person}{rgb}{0.859,0.078,0.234}
\definecolor{color_roadline}{rgb}{0.613,0.914,0.195}
\definecolor{color_sidewalk}{rgb}{0.953,0.137,0.906}

\definecolor{ellisred}{rgb}{0.87,0.44,0.38} %
\definecolor{ellisgreen}{rgb}{0.69,0.90,0.52} %
\definecolor{elliscyan}{rgb}{0.29,0.77,0.74} %
\definecolor{ellisorange}{rgb}{0.89,0.55,0.28} %
\definecolor{ellisblue}{rgb}{0.41,0.61,0.86} %

\newcommand{\ellisred}[1]{\noindent{\color{ellisred}{#1}}}

\newcommand{\ellisblue}[1]{\noindent{\color{ellisblue}{#1}}}

\definecolor{C0}{HTML}{1F77B4}
\definecolor{C1}{HTML}{ff7f0e}
\definecolor{C2}{HTML}{2ca02c}
\definecolor{C3}{HTML}{d62728}
\definecolor{C4}{HTML}{9467bd}
\definecolor{C5}{HTML}{8c564b}
\definecolor{C6}{HTML}{e377c2}
\definecolor{C7}{HTML}{7f7f7f}
\definecolor{C8}{HTML}{bcbd22}
\definecolor{C9}{HTML}{17becf}

\definecolor{Cx0}{HTML}{4e79a7}
\definecolor{Cx1}{HTML}{f28e2b}
\definecolor{Cx2}{HTML}{e15759}
\definecolor{Cx3}{HTML}{76b7b2}
\definecolor{Cx4}{HTML}{59a14f}
\definecolor{Cx5}{HTML}{edc948}
\definecolor{Cx6}{HTML}{b07aa1}
\definecolor{Cx7}{HTML}{ff9da7}
\definecolor{Cx8}{HTML}{9c755f}
\definecolor{Cx9}{HTML}{bab0ac}

\newcommand{\name}{SLEDGE\xspace}

\usepackage{pifont}%
\newcommand{\cmark}{\ding{51}}%
\newcommand{\xmark}{\ding{55}}%

\newcommand{\pmsd}[1]{{{\scriptsize $\pm$ #1}}}

\newcommand\blfootnote[1]{%
  \begingroup
  \renewcommand\thefootnote{}\footnote{#1}%
  \addtocounter{footnote}{-1}%
  \endgroup
}

\begin{document}

\title{SLEDGE: Synthesizing Driving Environments with Generative Models and Rule-Based Traffic} 

\titlerunning{\name}

\author{Kashyap Chitta* \quad
Daniel Dauner* \quad
Andreas Geiger}
\authorrunning{K. Chitta et al.}
\institute{University of Tübingen \quad \quad
Tübingen AI Center\\
\href{https://github.com/autonomousvision/sledge}{https://github.com/autonomousvision/sledge}}

\maketitle

\blfootnote{*equal contribution}
\vspace{-0.9cm}

\begin{abstract}
    SLEDGE is the first generative simulator for vehicle motion planning trained on real-world driving logs. Its core component is a learned model that is able to generate agent bounding boxes and lane graphs. The model's outputs serve as an initial state for rule-based traffic simulation. The unique properties of the entities to be generated for SLEDGE, such as their connectivity and variable count per scene, render the naive application of most modern generative models to this task non-trivial. Therefore, together with a systematic study of existing lane graph representations, we introduce a novel raster-to-vector autoencoder. It encodes agents and the lane graph into distinct channels in a rasterized latent map. This facilitates both lane-conditioned agent generation and combined generation of lanes and agents with a Diffusion Transformer. Using generated entities in SLEDGE enables greater control over the simulation, e.g. upsampling turns or increasing traffic density. Further, SLEDGE can support 500m long routes, a capability not found in existing data-driven simulators like nuPlan. It presents new challenges for planning algorithms, evidenced by failure rates of over 40\% for PDM, the winner of the 2023 nuPlan challenge, when tested on hard routes and dense traffic generated by our model. Compared to nuPlan, SLEDGE requires 500$\times$ less storage to set up (<4 GB), making it a more accessible option and helping with democratizing future research in this field.
    
\end{abstract}

\section{Introduction}
\label{sec:intro}

While recent breakthroughs in generative AI have revolutionized natural image synthesis~\cite{Esser2024ARXIV,Brooks2024}, generative models are yet to find widespread adoption in autonomous driving. In contrast to the regular pixel grid of images, self-driving planners typically require abstract bird's eye view (BEV) representations as input which characterize the most important scene elements (\eg, lanes, traffic lights, static and dynamic objects) in a compact, vectorized format (see \figref{fig:raster}). These representations are a key component of data-driven simulators which are necessary for rigorous evaluation of planners~\cite{Dauner2023CORL,Zhai2023ARXIV,Li2023ARXIV,Contributors2024}. However, learning a generative model on such irregular vectorized representations is challenging.

Consequently, many existing data-driven simulators~\cite{Karnchanachari2024ICRA,Gulino2023NIPS} are initialized by simply replaying logs of abstract representations. They extract the local lane graph from a High Definition (HD) map and object bounding boxes from pre-recorded annotation logs. Planning algorithms can be tested in scenarios with restricted routes and durations ($\sim$15 seconds long), to ensure that the environment during simulation is covered in the recording.
Moreover, to provide sufficient diversity among routes for comprehensive testing, these simulators require huge databases, \eg, nuPlan~\cite{Karnchanachari2024ICRA} consists of 1300 hours of driving logs which require over 2 TB of storage. Such high resource requirements heighten the barrier for entry into the field of vehicle motion planning.

In this paper, we study the generation of simulation-ready abstract representations of driving scenes. Using generative models as an alternative to log replays has the potential for significant compression~\cite{Santurkar2017ARXIV}. However, the unique characteristics of abstract representations in driving scenes pose new challenges to modeling them, e.g., they require accurate topological connectivity, variable entity counts, and precise modeling of geometry (e.g., parallel lines). Due to these characteristics, generative models that operate on uniformly sized representations (e.g., images) are incompatible with our data-driven simulation task.

\begin{figure}[t!]
    \centering
    \includegraphics[width=\linewidth]{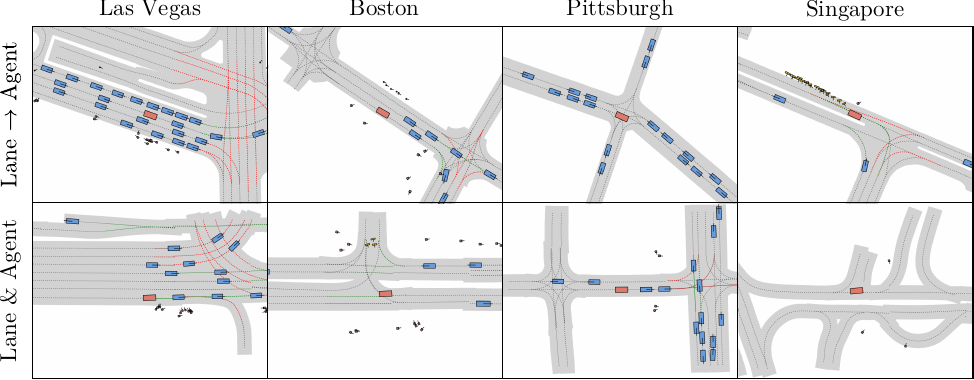} \\
    \caption{\textbf{SLEDGE.} We show state snapshots of simulation environments generated by our approach in 4 cities, with the \textcolor{gray}{lanes}, \ellisred{ego-vehicle}, \ellisblue{other vehicles}, \textcolor{Cx6}{pedestrians}, \textcolor{Cx5}{obstacles}, and \textcolor{C3}{traffic} \textcolor{C2}{lights}. Our supplementary video visualizes clips with more examples.
    }
    \label{fig:sim_1}
    \vspace{-0.8cm}
\end{figure}

To tackle these complexities, we first perform a systematic study of the existing representations of lane graphs used in autonomous driving. We then propose a novel representation based on a raster-to-vector autoencoder. It represents a driving scene with a fixed-size BEV rasterized latent map (RLM). We learn to generate these RLMs with a Diffusion Transformer (DiT)~\cite{Peebles2023ICCV}. Our model generates high-fidelity results enabling both lane-conditioned agent generation or joint lane and agent generation within a single flexible and scalable framework (\figref{fig:sim_1}).
Independent to its use of generative models, \name also provides the previously missing functionality of simulating only agents within a certain radius of the ego-vehicle. By doing so, we can test on routes that are significantly longer than those currently used for evaluating planners. We find previously ignored failure modes of the state-of-the-art PDM-Closed planner~\cite{Dauner2023CORL}, which is unable to complete 25-50\% of our new 500m long test routes despite having failure rates below 10\% on existing benchmarks. 

\boldparagraph{Contributions}
(1) We formalize the task of abstract scene generation for autonomous driving with a challenging benchmark and corresponding metrics.
(2) We perform a systematic exploration of modern generative models (with various architectures and representations) and propose a novel latent diffusion model for synthesizing abstract driving scenes that largely outperforms other baselines.
(3) We present a simulation framework, \name, which is nearly 3 orders of magnitude more storage efficient than nuPlan yet enables more rigorous testing of planning algorithms with long-horizon simulations using rule-based traffic.

\section{Related Work}
\label{sec:related}

\boldparagraph{Diffusion Models} Best known for their success in generative modeling of images~\cite{Rombach2022CVPR,Esser2024ARXIV} and video~\cite{Blattmann2023ARXIV,Brooks2024,Yang2024ARXIV}, diffusion models have recently found widespread adoption in diverse domains, including point clouds~\cite{Luo2021CVPR,Zeng2022NIPS,Nichol2022ARXIV}, floor plans~\cite{Shabani2023CVPR,Chen2023NIPSa}, molecules~\cite{Xu2023ICML}, robot policies~\cite{Chi2023RSS}, traffic patterns~\cite{Zhong2023ICRA}, and many others. Scenario Diffusion~\cite{Pronovost2023NIPS}, a pioneering approach in generating vehicles conditioned on HD maps with diffusion, is the closest existing approach to ours. This method uses latent diffusion with a raster-based vehicle decoder unlike our proposed transformer decoder head. Importantly, we offer significantly increased capabilities: generation of lane graphs, support for pedestrians, obstacles, and traffic lights, as well as long-horizon simulation environments with reactive agents.

\boldparagraph{Generating Lane Graphs} Lane graphs, the most important component of HD maps, are well-studied. They are often constructed through an offline mapping process often involving human annotators~\cite{Elhousni2020ARXIV}. However, a surge of recent work on predicting lane graphs from sensor data~\cite{Liao2023ICLR,Li2024ICLR} has sparked interest on generative modeling of these graphs. The first and only existing study on this task, HDMapGen~\cite{Mi2021CVPR}, proposes an autoregressive approach for generating lane graphs node-by-node~\cite{Chu2019ICCV}. Our experiments show that this achieves reasonable results, but is unable to match the high quality and scalability offered by our model. Unlike HDMapGen, our approach jointly generates agents with lanes, and efficiently generates all elements in parallel. A concurrent project, DriveSceneGen~\cite{Sun2023ARXIV}, generates lanes and vehicles with image-space diffusion. Our approach covers agent types beyond vehicles, uses less heuristics, and is more efficient. %
Additionally, we show the successful integration of our model into a simulator.%

\boldparagraph{Data-driven Simulation} Developing an autonomous driving system necessitates rigorous testing which is costly and risky if conducted in the real world. Driving simulators are an alternative~\cite{Wymann2015, Dosovitskiy2017CORL, Karnchanachari2024ICRA}. However, simulators face challenges in ensuring realism while initializing traffic scenes, simulating traffic. Instead, data-driven simulators address these challenges by replaying traffic scenes from real-world recordings~\cite{Althoff2017IV, Karnchanachari2024ICRA, Li2022PAMI,Gulino2023NIPS}. The simulator can mine specific situations or even optimize the initial parameters for safety-critical scenarios~\cite{Ding2021RAL, Hanselmann2022ECCV, Wang2021CVPRb}. Leveraging modern generative models, we take a step further than existing frameworks and learn the underlying distribution of the real-world data. %

\section{Method}
\label{sec:method}

Our goal is to design a driving scene synthesis framework that can be trained using real-world driving logs and incorporated into SLEDGE, our generative simulator with rule-based traffic. We base this framework on LDMs~\cite{Rombach2022CVPR} as: (1) latent diffusion shows excellent training stability and scalability with compute. (2) One can easily construct a fixed-size latent space for diffusion that can be mapped to the variable sized set representation for simulation (\secref{sec:method_vector}) using detection-based transformer architectures~\cite{Carion2020ECCV,Liao2023ICLR}. Our LDM is trained in two stages: an autoencoder (\secref{sec:method_rvae}) followed by a diffusion model (\secref{sec:method_dit}). We detail the simulation of scenes generated by the LDM in \secref{sec:method_sim}.

\subsection{nuPlan Vector Representation}
\label{sec:method_vector}

We combine sets of entities to represent scenes in the default nuPlan format. %

\boldparagraph{Lanes} Our focus is on the generation of lanes, the central element of HD maps used in data-driven simulation. Each lane $\bL \in \nR^{20 \times 2}$ is geometrically represented by a polyline, \ie, a fixed set of 20 bird’s eye view (BEV) points. These are bounded by two endpoints and form the lane centerline, connected along the driving direction. A lane may share endpoint(s) with predecessor and successor lanes. This information is encoded in an adjacency matrix $\bA \in \nR^{N \times N}$, where N is the number of lanes in a certain field of view (FOV). The set of all lane polylines $\cL$ form the lane graph of the local map $\cM = \{ \cL, \bA \}$.

\boldparagraph{Traffic Lights} We then augment the lane graph with polylines representing traffic lights. These share the same ${20 \times 2}$ format as lanes, and come in two types (red and green). The set of red polylines ($\cR$) contains the lane regions that are currently not traversable due to a red traffic light. The set of green polylines ($\cG$), on the other hand, contains entities indicating lane segments where the road is safe to proceed along due to the presence of a green traffic signal.

\boldparagraph{Agents} Further, we expand the scene representation using oriented bounding boxes for the agents. Each bounding box is defined by a 2D center position, heading, 2D extent and optional speed. We consider three types of agent sets: pedestrians ($\cP$), vehicles ($\cV$) and static objects ($\cO$). Pedestrians and vehicle boxes are assigned a speed attribute, whereas static objects are not. 

\boldparagraph{Ego Velocity} Finally, initializing a simulation requires the BEV ego velocity $\bv \in \nR^2$. Overall, we denote the scene state as $\cS = \{\cM, \cR, \cG, \cP, \cV, \cO, \bv\}$. %

\subsection{Raster-to-Vector Autoencoder}
\label{sec:method_rvae}

Each entity type in $\cS$ is unique. To maintain overall scene consistency, we would like to model them with a single architecture, instead of creating several independent entity-specific generative models. Furthermore, most existing tools in the literature have been developed and optimized for 2D input domains~\cite{Rombach2022CVPR}. To this end, we propose the raster-to-vector autoencoder (RVAE) which unifies all entity types in $\cS$ into a compact, shared 2D representation well-suited for diffusion modeling. The autoencoder is inspired by work in object detection~\cite{Carion2020ECCV} and online mapping for autonomous driving~\cite{Liao2023ICLR}.

\begin{figure}[t!]
    \centering
    \includegraphics[width=0.95\linewidth]{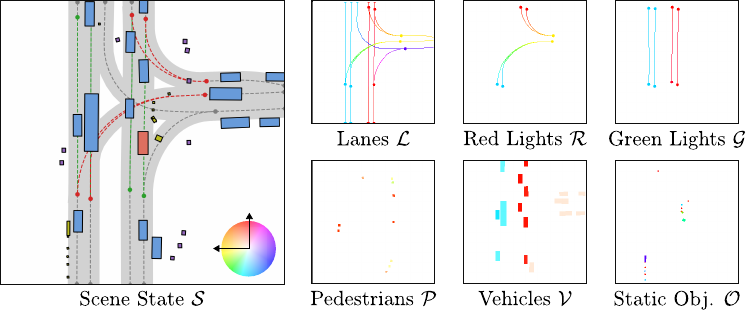} \\
    \caption{\textbf{Rasterized State Image (RSI).} We encode $\cS$ into a 12-channel image, with 2 channels per entity type. We visualize these encodings as optical flow fields.
    }
    \label{fig:raster}
    \vspace{-0.5cm}
\end{figure}

\boldparagraph{Rasterization} We first define a function $\rho \colon \cS \rightarrow \bI$ that encodes the scene state into a rasterized state image (RSI) $\bI \in \nR^{W \times H \times 12}$. Our design of $\rho$ is motivated by (and closely resembles) techniques used in motion planners~\cite{Bansal2019RSS,Renz2022CORL}. As shown in \figref{fig:raster}, $\rho$ maps the three polyline entity types ($\cL, \cR, \cG$) and three bounding box entity types ($\cP, \cV, \cO$) to image pixel locations, assigning 2 channels to each entity type which encode all their attributes. For polylines, we use a 2D directional vector $\Delta = [dx, dy]$, which points from any point to its successor, indicating the presence of a polyline traversing a specific pixel. A background value (\ie, $[0, 0]$) is assigned to other regions. For a bounding box type entity, we rasterize it in BEV according to its position, extent, and heading. For dynamic bounding boxes ($\cP, \cV$), the values in the two channels within the box region represent the entity's 2D velocity. For static obstacles, we fill the rasterized region with the obstacle's orientation vector. We use a square field of view centered at and oriented as per the ego vehicle's pose for rasterization. The ego velocity $\bv$ is encoded at the origin of $\bI$ as an extra rasterized vehicle in $\cV$.

\begin{figure}[t!]
    \centering
    \includegraphics[width=\linewidth]{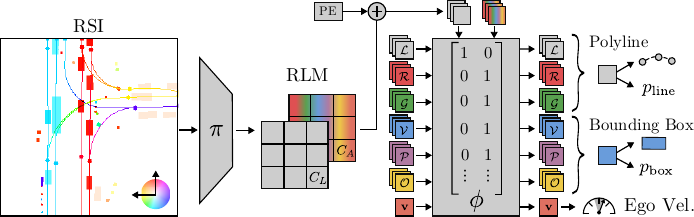} \\
    \caption{\textbf{Raster-to-Vector Autoencoder (RVAE).} We represent scenes with a rasterized latent map (RLM) consisting of two channel groups. The `Lanes' group is decoded into lane segments and the `Agents' group into all other scene entities, via a transformer decoder with attention masking. The autoencoder is trained to predict polylines, bounding boxes, and the ego velocity in a simulation-compatible format.
    }
    \label{fig:rvae}
    \vspace{-0.3cm}
\end{figure}

\boldparagraph{Vectorization} This step is necessary to decode the unified RSI representation $\bI$ back into the per-entity attributes (\eg, polylines, bounding boxes). We use a learned vectorization pipeline consisting of a raster encoder $\pi$ and vector decoder head $\phi$. The ResNet-50~\cite{He2016CVPR} encoder takes the RSI and outputs a rasterized latent map (RLM) $\bM = \pi (\bI)$ of shape $W' \times H' \times C$, where $H' = H / 2^d$ and $W' = W / 2^d$ for a downsampling factor $d \in \nN$. $C$ is a chosen channel dimension.  In practice, the RLMs we use are compact, $H' = W' = 8$ and $C=64$. Additionally, as shown in \figref{fig:rvae}, we split the RLM's channels into 2 groups, $C = C_L + C_A$ for the lanes and agents respectively. For each group, we tokenize the latent vectors spatially with $1 \times 1$ patches, resulting in $W' \times H'$ lane tokens and $W' \times H'$ agent tokens. Following the DETR~\cite{Carion2020ECCV} paradigm, we use these tokens as keys and values for our transformer decoder $\phi$. The decoder uses a fixed number of learnable queries of each entity type, which we cap to a maximum count per entity, based on statistics from our dataset. The final decoder layer is unique per entity type and outputs the attributes specific to that entity, \eg a $20\times2$ set of point coordinates for a polyline, or a 6-dimensional descriptor (2D position, orientation, 2D extent, and speed) for a bounding box. It also outputs an existence attribute $p\in [0,1]$ for both polylines ($p_{line}$) and bounding boxes ($p_{box}$), which is used to handle variable counts of ground truth entities with a fixed number of queries.

\boldparagraph{Channel Group Masking} The motivation behind our design with two token groups is to enable agent generation conditioned on known lanes. To this end, the tokens for agents should contain no information about lanes. We implement a binary mask in the cross-attention mechanism of $\phi$ to achieve this. Specifically, queries for lanes $\cL$ are prevented from attending to the keys and values of the agents tokens, and all other queries (\ie,  $\cR, \cG, \cP, \cV, \cO, \bv$) cannot attend to the lanes tokens. Our experiments show the effectiveness of this approach.

\boldparagraph{Training} The autoencoder is optimized using both reconstruction and existence losses, and a KL divergence loss on the RLM. For reconstruction, we first match generated and ground truth entities using the Hungarian algorithm, as in~\cite{Carion2020ECCV}. We use a matching score based on the L1 error of the entity's position attributes. We then use the L1 error summed over all attributes and averaged over all matches as the training loss. For the existence variable, we use a binary cross entropy loss based on whether the query was matched to a ground truth entity. %

\subsection{Diffusion Transformer}
\label{sec:method_dit}

We obtain RLMs $\bM$ for each training example via the frozen, pretrained encoder $\pi$ and use them to train a diffusion model $\delta$ with the DDPM algorithm~\cite{Ho2020ARXIV}. 

\boldparagraph{Training} For each scene, we sample a noise scaling factor $\sigma$ from a log-normal distribution and create a noisy sample $\hat \bM = \bM + \sigma \boldsymbol{\cE}$, with $\boldsymbol{\cE}$ sampled from a standard normal distribution of the same shape as $\bM$. We model $\delta(\hat \bM; \bc, \sigma)$ as a Transformer~\cite{Vaswani2017NIPS}, following DiT~\cite{Peebles2023ICCV}, where $\bc$ is a conditioning vector. We choose $\bc$ to be a one-hot label indicating the city to which the example belongs. This conditioning resolves ambiguities between locations (e.g. right- and left-hand driving in the US and Singapore).
However, it is possible to adapt our framework to other conditioning, \eg, images, text descriptions, or learned clusters based on the autoencoder features. The DiT architecture is simple, scalable, and free from down- or upsampling operations, making it compatible with RLMs of any spatial resolution. It applies a series of self-attention blocks to the tokenized input $\hat \bM$, with the conditioning on $\bc$ and $\sigma$ implemented using AdaLN-Zero~\cite{Peebles2023ICCV}. We optimize the L2 reconstruction loss between $\boldsymbol{\cE}$ and $\delta(\hat \bM; \bc, \sigma)$.

\boldparagraph{Generation} During inference, we begin with an initial noisy sample $\hat \bM \sim \mathcal N \left( 0, \sigma_\textrm{\small max}^2 \mathbf{I} \right)$, which undergoes iterative refinement from $\sigma = \sigma_\textrm{\small max}$ to $\sigma = 0$ based on the reverse PDE defined via $\delta$~\cite{Ho2020ARXIV}. We then use the trained vector decoder $\phi$ from \secref{sec:method_rvae} to predict $\cL, \cR, \cG, \cP, \cV, \cO,$ and $\bv$. Only entities with an existence probability above a threshold $\tau$ are retained, and for overlapping bounding boxes, those with the highest existence probability are kept. The process of recovering the full scene state $\cS$ further involves extracting the adjacency matrix $\bA$ of the lane graph, which is not an explicit output of our model. We do this by simply matching lanes whose start and endpoints lie within a range of 1.5 meters with orientations differing by less than 60 degrees, which we found to be robust in practice given our highly accurate lane polyline predictions.

\begin{figure}[t!]
    \centering
    \includegraphics[width=1.0\linewidth]{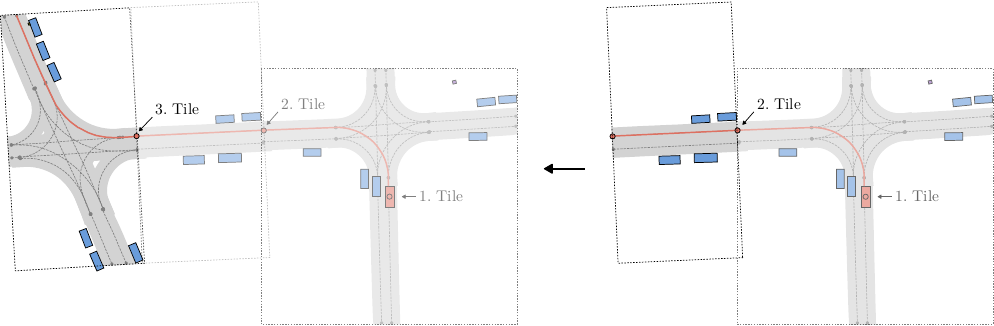} \\
    \caption{\textbf{Route Extrapolation by Inpainting.} We show an example scenario generated by our DiT, where we iteratively sample poses along a route, warp the previous tile's RSI to this pose, and generate a new tile conditioned on the warped RSI.  
    }
    \label{fig:extrapolation}
    \vspace{-0.3cm}
\end{figure}

\boldparagraph{Conditional Generation via Inpainting} Diffusion models excel at inpainting, even without explicit training for this task~\cite{Lugmayr2022CVPR}. Specifically, by executing the denoising process on only a subset of the tokens in the noisy sample $\hat \bM$, with the remaining tokens extracted from a known and encoded scene, we can inpaint RLMs. We use this capability to perform two tasks: (1) lane conditioned agent generation and (2) route extrapolation. Lane conditioned agent generation involves encoding all lane tokens from a known map, and denoising all agent tokens. Route extrapolation, as illustrated in \figref{fig:extrapolation}, involves encoding a subset of tokens within a known spatial region to denoise the unknown region. Specifically, we can iteratively sample poses along a generated route, warp the previously generated scene's RSI to the new pose with an affine transformation, and use a known region as conditioning for completing a newly created tile. We provide implementation details in the supplementary material.

\subsection{SLEDGE Simulation Environments}
\label{sec:method_sim}

Finally, we initialize a reactive simulation in SLEDGE using the generated initial scene state $\cS$. In the following, we provide an overview of the steps involved.

\boldparagraph{Hard Routes and Traffic} To evaluate a planner in ambiguous situations, we must specify the driver intention, \eg whether to turn left or right at an intersection. Existing replay-based simulators offer limited controllability over this, since they are unable to extract agents to simulate if the planner diverges significantly from the route followed by the human driver from the log recording. However, for generated scenes, we can extract multiple valid routes from the lane graph, \eg, we define `hard' routes by selecting the route with the highest number of turns. In addition, our approach also provides a degree of control over traffic density. We define a `hard' traffic setting by generating multiple valid traffic configurations along the desired route, and selecting the configuration with the largest number of generated agents. Our experiments show that these `hard' settings provide new challenges to the state of the art for planning.

\boldparagraph{Behavior Simulation} We simulate non-ego vehicles in SLEDGE by projecting each to the center of a lane based on proximity and heading, from which it then laterally follows the lane centerlines. For longitudinal control, we use a simple policy called the Intelligent Driver Model (IDM)~\cite{Treiber2000}. Upon choosing a connected lane sequence as a driving path, IDM calculates a longitudinal trajectory, iteratively adjusting acceleration based on current position, velocity, and distance to the leading vehicle. For pedestrians, we assume a constant velocity and heading while unrolling the simulation. Traffic lights are hard-coded to change states every 15 seconds. While these choices are simplistic, they are in line with the capabilities of today's best data-driven simulators~\cite{Karnchanachari2024ICRA,Gulino2023NIPS}. SLEDGE is not incompatible by design with other types of policies for vehicle and pedestrian simulation, but we leave this exploration to future work.

\begin{figure}[t!]
    \centering
    \includegraphics[width=\linewidth]{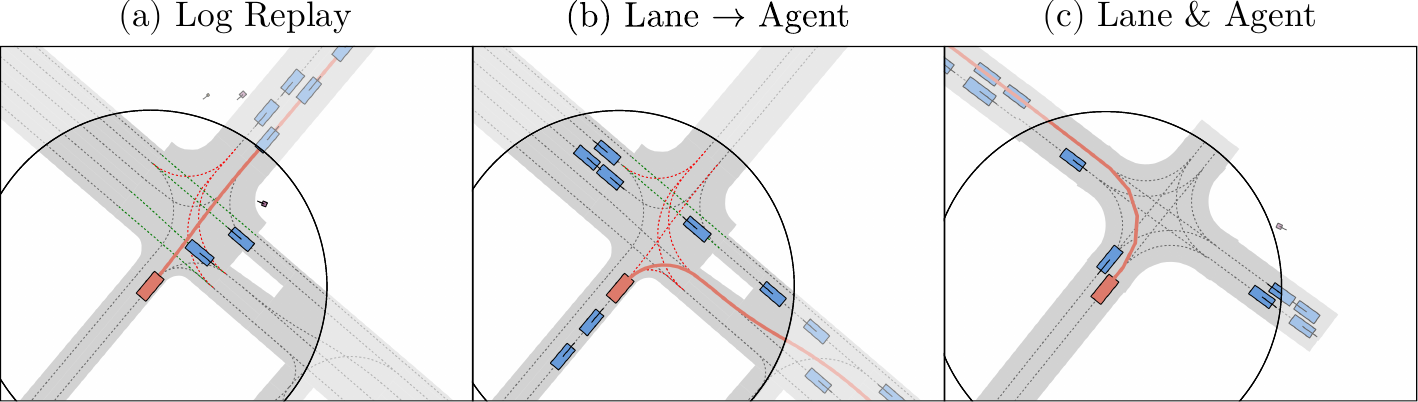} \\
    \caption{\textbf{Long Route Simulation.} SLEDGE supports (a) replayed scenarios, (b) lane-conditioned agent generation, and (c) joint lane and agent generation. Importantly, we enable testing on arbitrarily long routes by dynamically simulating agents near the ego vehicle while keeping the state of distant agents fixed.
    }
    \label{fig:teaser}
    \vspace{-0.3cm}
\end{figure}

\boldparagraph{Simulation Radius} By default, existing data-driven simulators like nuPlan simulate all the initialized agents at all timesteps. This severely limits the scalability of these simulators to long simulation horizons or large scenes. We propose a simple modification for SLEDGE wherein we only simulate agents at a distance below $\alpha$=64m to the ego vehicle at a given timestep, while holding the state of all other agents to be constant (\figref{fig:teaser}). We demonstrate the scalability this provides by conducting experiments on simulations that are 10$\times$ longer than those in nuPlan, \ie, up to 150 seconds as opposed to nuPlan's 15 seconds.

\section{Experiments}
\label{sec:result}

We now present our experimental results. (1) We demonstrate the suitability of the RLM as a lane graph representation. (2) We benchmark a series of models for lane graph generation. (3) We showcase the effectiveness of SLEDGE environments for evaluating planners. For all experiments, we present concise descriptions of baselines, metrics and implementations in the main paper. Additional details can be found in the supplementary material.

\boldparagraph{Dataset} We use nuPlan~\cite{Karnchanachari2024ICRA}, the largest publicly available dataset for vehicle motion planning. It comprises 1300 hours of logs from 4 cities. We sample 450k train and 50k validation frames from these logs with sampling intervals of 30s, 1s, 2s and 2s between frames for Las Vegas, Boston, Pittsburgh and Singapore respectively, in order to obtain a balanced distribution while achieving high map coverage. Each city offers unique challenges for generative modeling, \eg Las Vegas has large and dense intersections, while Singapore involves left-hand traffic. Each frame is limited to a 64m$\times$64m FOV centered at the ego vehicle. 

\boldparagraph{Implementation} We consider a ResNet-50 for the raster encoder $\pi$ and a transformer decoder with three layers for the vector decoder $\phi$ of the RVAE. For the generative tasks, we apply the DiT-L and DiT-XL (138M vs. 487M params) variants with a $1 \times 1$ patch size and DDPM noise scheduling as in~\cite{Rombach2022CVPR}. %

\subsection{Lane Graph Representations} 
\label{sec:result_rep}

In our first experiment, we evaluate various representations based on their ability to reconstruct the complete directed lane graph $\cM = \{ \cL, \bA \}$.

\boldparagraph{Baselines} We consider three baseline representations. \textbf{(1) RSI:} we use the skan library~\cite{NunezIglesias2018} to extract vector polylines from the $256\times256\times2$ RSI representation of the lane graph. Skan is a highly optimized image processing pipeline for graph extraction (details and visualizations in supplementary). \textbf{(2) RLM w/o mask:} we train the RVAE without the channel group masking proposed in \secref{sec:method_rvae}, which entangles information about agents and lanes into a single $8\times8\times64$ latent map instead of an $8\times8\times32$ tensor for the lane graph and an independent $8\times8\times32$ tensor for agents. \textbf{(3) Vector:} As an upper bound, we additionally compute our metrics for the respresentation used as target labels for the autoencoder during training. This is a set of polylines shaped $N\times20\times2$, where we cap $N$ at 30 in our experiments. Note that the ground truth for evaluation in this experiment uses all polylines in the scene (which is sometimes greater than 30).

\boldparagraph{Metrics} Our metrics are adapted from the street map extraction literature~\cite{He2022WACV}. We use three base metrics, which all operate on point sets sampled along graphs at a resolution of one point every 1.5m. \textbf{(1) F1:} measures the harmonic mean of the precision and recall, which are estimated using Hungarian matching between point sets with a distance threshold of 1.5m. Intuitively, this penalizes large structural errors, while ignoring small positional offsets. \textbf{(2) Lateral L2 (Lat.):} averaged over all true positive matched points, measures the lateral offset of each such point from its nearest ground truth lane centerline. In contrast to F1, it penalizes positional errors, while ignoring structurally incorrect and unmatched lanes. \textbf{(3) Chamfer:} averaged over all points in two point sets, this is the distance of each point to the closest in the other set. It requires both precise structure and details. These three base metrics are further applied in two settings. \textbf{(1) GEO} uses point sets sampled from the complete graph, making it independent of the predicted adjacency matrix $\bA$. On the other hand, \textbf{(2) TOPO} uses $\bA$ to extract sets of fully-connected sub-graphs corresponding to every tenth node of the graph (\ie, every 15 meters). The base metrics are computed on these sub-graphs and averaged. Errors in $\bA$ can lead to large missing sections of such sub-graphs, making TOPO suitable for evaluating connectivity. 

\begin{table}[t]
    \caption{\textbf{64m$\times$64m Lane Graph Reconstruction.} We show the F1 score, lateral displacement and Chamfer distance for graphs extracted from each representation. We additionally include qualitative results (more in supplementary material). The RSI struggles with nearly overlapping segments at the beginning of forks (FNs) and overlaps at intersections (FPs). The RLM closes the gap towards the upper bound (vector).
    }
	\centering
    \resizebox{\linewidth}{!}{
    \setlength{\tabcolsep}{6pt}
	\begin{tabular}{l|ccc|c|c|c|c|c|c}
        \toprule
		\multirow{2}{*}{\textbf{Rep.}} & \multirow{2}{*}{\textbf{Fixed?}} & \multirow{2}{*}{\textbf{Split?}} & \textbf{Size} & \multicolumn{3}{c|}{\textbf{GEO}} & \multicolumn{3}{c}{\textbf{TOPO}} \\
        \cmidrule{5-10}
        & & & \textbf{(KB)} & {F1} $\uparrow$ & {Lat.} $\downarrow$ & {Ch.} $\downarrow$ & {F1} $\uparrow$ & {Lat.} $\downarrow$ & {Ch.} $\downarrow$ \\
        \midrule
        RSI & \cmark & \cmark & 524.3 & 0.933 & {0.133} & 0.423 & 0.851 & 0.438 & 64.824 \\
        \midrule
        \multirow{2}{*}{RLM} & \multirow{2}{*}{\cmark} & \xmark & 16.0 & {0.981} & 0.161 & {0.399} & {0.945} & {0.282} & {20.096} \\
        & & \cmark & 8.0 & 0.980 & 0.164 & 0.411 & 0.944 & 0.288 & 20.624 \\
        \midrule
        \textit{Vector} & \xmark & \cmark & \textit{4.8} & \textit{0.997} & \textit{0.005} & \textit{0.070} & \textit{0.990} & \textit{0.010} & \textit{4.174} \\
        \bottomrule
	\end{tabular}
    }
    \begin{center}\includegraphics[width=0.95\linewidth]{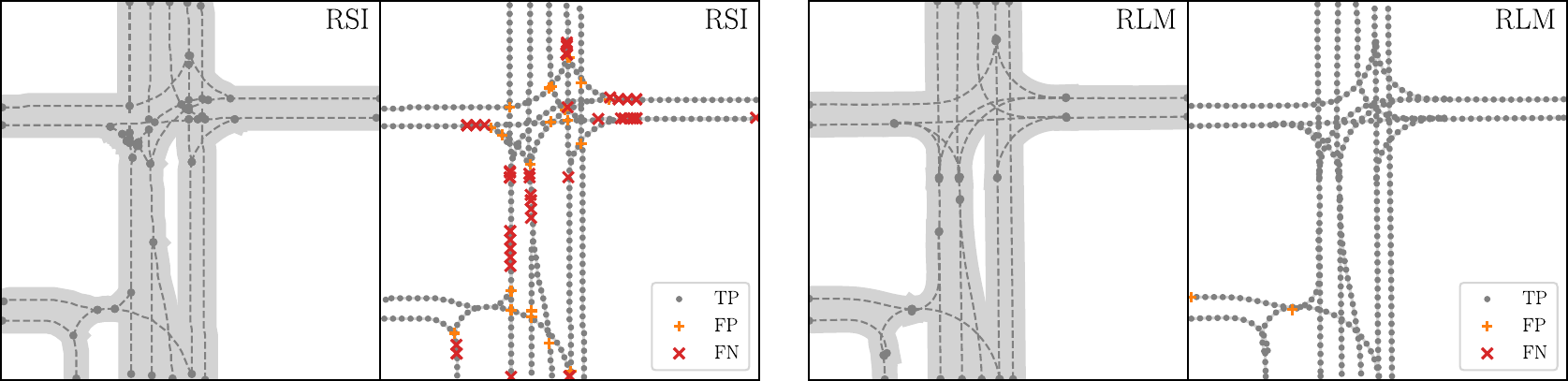}
    \end{center}
	\label{tab:rep}
	\vspace{-0.3cm}
\end{table}

\boldparagraph{Results} As shown in \tabref{tab:rep}, the RSI is unable to match the RLM on all TOPO and both the F1 and Chamfer GEO metrics. Despite the significantly larger representation size of 524 KB per 64m$\times$64m scene, the reliance on heuristics to convert this representation back to its constituent entities serves as a major limiting factor for the RSI. Qualitatively, we observe that the key issue is dense groups of lines that nearly overlap at forks or cross over in intersections. For the 2 variants of RLM, we obtain similar reconstruction quality, largely closing the gap towards the upper bound vector representation. Importantly, the proposed channel group masking for the RLM (\ie, `Split') disentangles the lane graph from the agents with no impact on the graph reconstruction fidelity. Finally, as we cap the maximum number of polylines to 30, we observe a very minor error rate (<1\% drop in F1) in the upper bound vector representation, corresponding to the negligible fraction of scenes with over 30 lanes. However, the size of the vector representation varies significantly per scene, ranging from 2 to 30 lanes. The RLM achieves an ideal balance of high quality with a fixed size.

\begin{table*}[t]
    \caption{\textbf{64m$\times$64m Lane Graph Generation.} We show the route lengths, precision and recall (in \%), and various Frechet distances between samples from the validation set and model outputs. We additionally include qualititve results (more examples in supplementary material). Latent diffusion combining DiT with an RLM obtains the best results. *Trained with $\sim$6$\times$ more compute than others, which already converge at the lower compute budget.
    }
	\centering
    \setlength{\tabcolsep}{5pt}
    \resizebox{\linewidth}{!}{
	\begin{tabular}{cc|c|rr|rr|rrrr}
        \toprule
		\multirow{2}{*}{\textbf{Arch.}} & \multirow{2}{*}{\textbf{Repr.}} & \textbf{Route} & \multicolumn{2}{c|}{\textbf{Prec. (CNN)} $\uparrow$} & \multicolumn{2}{c|}{\textbf{Recall (CNN)} $\uparrow$} & \multicolumn{4}{c}{\textbf{Frechet (Urban Planning)} $\downarrow$} \\
        \cmidrule{4-11}
        & & \textbf{Length} $\uparrow$ & ImNet & RVEnc & ImNet & RVEnc & Conne. & Densi. & Reach & Conve. \\
		\midrule
        {VAE} & RSI & {2.68} \pmsd {3.66} & 26.58 & 0.00 & 8.70 & 0.16 & 9.45 & 0.99 & 2.86 & 13.06 \\
        RVAE & Vector & {23.79} \pmsd {9.96} & 10.41 & 4.56 & 16.28 & 8.14 & 15.63 & 12.57 & 3.08 & 17.72 \\
        HDMapGen & Vector & {28.17} \pmsd {14.81} & 19.10 & 7.48 & 17.17 & 12.45 & 7.02 & 3.03 & 2.49 & 18.10 \\
        \midrule
		\multirow{2}{*}{DiT-L} & RSI & {24.78} \pmsd {10.38} & 20.36 & 19.20 & 21.49 & 5.94 & 6.11 & 15.33 & 1.90 & 3.95 \\
        & RLM & {32.51} \pmsd {9.93} & 33.25 & 63.99 & 36.24 & 61.60 & 2.35 & 3.52 & 0.88 & 3.10 \\
        \midrule
        DiT-XL* & RLM & \textbf{{35.37} \pmsd {10.28}} & \textbf{42.05} & \textbf{78.07} & \textbf{42.32} & \textbf{72.63} & \textbf{0.27} & \textbf{2.47} & \textbf{0.20} & \textbf{0.47} \\
        \bottomrule
        \rule{0pt}{2ex} \\
	\end{tabular}
    }
    \includegraphics[width=\linewidth]{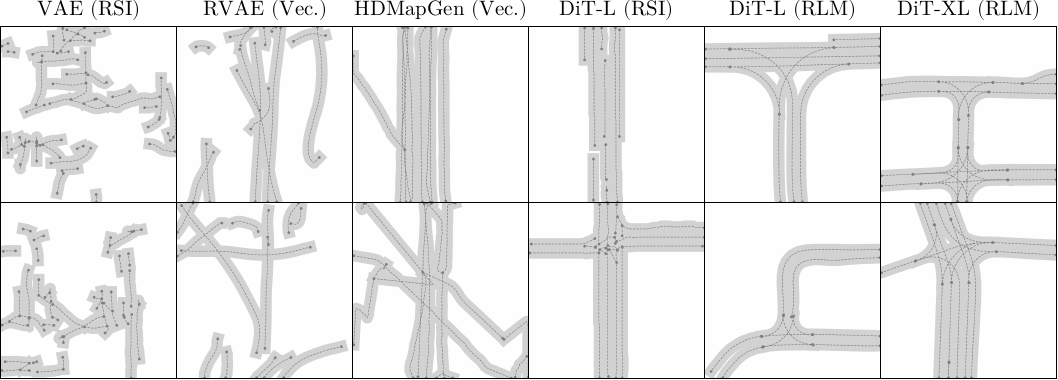}
	\label{tab:gen}
	\vspace{-0.7cm}
\end{table*}

\subsection{Lane Graph Generation} 
\label{sec:result_gen}

Next, we compare our proposed DiT to several generative models for lane graphs. 

\boldparagraph{Baselines} We select four diverse baselines. \textbf{(1) VAE:} we train a convolutional VAE with a 2D decoder head to generate RSIs. \textbf{(2) RVAE:} we sample from the decoder $\phi$ of our proposed autoencoder. \textbf{(3) HDMapGen:} an autoregressive hierarchical graph neural network for lane graph generation~\cite{Mi2021CVPR}, reimplemented for nuPlan. \textbf{(4) DiT (RSI):} similar to the concurrent DriveSceneGen~\cite{Sun2023ARXIV}, this is an image diffusion model that generates the RSI. %

\boldparagraph{Metrics} Lane graph generation does not have established evaluation protocols. Therefore, we use a comprehensive set of four metric types. \textbf{(1) Route Length} measures the mean and std of the longest valid ego vehicle route in the 64m$\times$64m FOV for 1k generated graphs. We assess sample quality with \textbf{(2) Precision} and coverage with \textbf{(3) Recall} based on the improved generative metrics from~\cite{Kynkaanniemi2019NIPS}. To obtain fixed-sized representations needed to estimate precision and recall with a nearest neighbour classifier, we rasterize the lane graphs and consider the penultimate feature vectors of a ResNet-50. We use two versions of these metrics with ResNets trained on ImageNet and the encoder $\pi$ of our autoencoder. The remaining metrics are \textbf{(4) Frechet} distances taken from~\cite{Chu2019ICCV,Mi2021CVPR}, based on graph features used in urban planning. They operate on nodes of the generated lane graphs with degree $\ne$ 2, which are referred to in~\cite{Mi2021CVPR} as \textbf{key points}. \textit{Connectivity:} this uses the degrees of all key points. \textit{Density:} the number of key points in the 64m$\times$64m FOV. \textit{Reach:} the number of valid paths found from key points to others. \textit{Convenience:} Lengths of all valid paths from all key points. The Frechet metrics are scaled by suitable powers of 10 for readability. Precision, Recall, and Frechet distances are measured using 50k generated and ground truth graphs.

\boldparagraph{Results} Our results are shown in \tabref{tab:gen}. All DiT variants generate more plausible layouts, with significantly better metrics than HDMapGen or the VAEs. For DiT-L, we observe higher coverage and visual fidelity when using the RLM representation instead of the RSI. In particular, the RLM-based models excel at creating coincident endpoints in intersections between connected lanes, which is crucial for smooth simulation. Scaling to DiT-XL provides further gains across all metrics, demonstrating the effectiveness of increasing the model capacity and compute budget. Since ImageNet features may be misaligned to BEV graphs, and the RVEnc features might favor RLM diffusion models, we focus on the urban planning Frechet metrics (in particular, Reach) in our subsequent experiments.

\begin{figure}[t!]
    \centering
    \includegraphics[width=\linewidth]{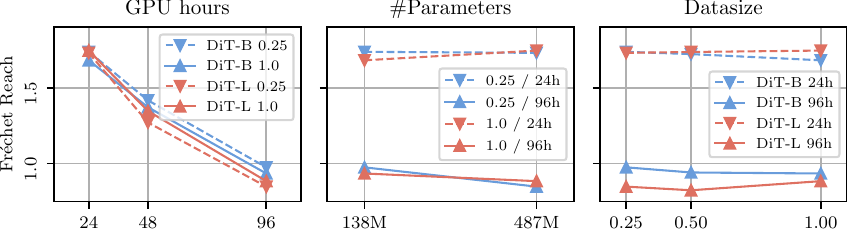} \\
    \caption{\textbf{Scaling.} The DiT's performance scales significantly with increased compute. For our task, dataset size is less crucial, with all settings performing similarly.
    }
    \label{fig:scaling}
    \vspace{-0.3cm}
\end{figure}

\boldparagraph{Scaling} We run a systematic analysis of scaling for our DiT with the RLM representation. We conduct a grid of experiments, considering \textbf{(1) 2 model sizes:} DiT-B and DiT-L, \textbf{(2) 3 dataset sizes:} 1$\times$, 0.5$\times$, and 0.25$\times$ our full dataset, and \textbf{(3) 3 compute budgets:} 24, 48 and 96 GPU hours. As shown in \figref{fig:scaling}, the performance scales significantly with increased compute. While it also scales with more parameters, more data does not have a large impact. This is possibly because data diversity is more valuable than scale, and all of our 3 datasizes have similar diversity. This scaling behavior shows the potential of further improvements for SLEDGE with more training resources.

\subsection{SLEDGE Simulation of PDM-Closed} 
\label{sec:result_sim}

In our final experiment, we use the inpainting capabilities of DiT-XL in SLEDGE to demonstrate its utility for testing vehicle motion planners.

\boldparagraph{Tasks and Settings} We consider the two inpainting-based tasks described in \secref{sec:method_dit}: lane conditioned agent generation (\textbf{Lane $\rightarrow$ Agent}) and joint lane and agent generation via route extrapolation (\textbf{Lane \& Agent}). For the Lane $\rightarrow$ Agent task, we consider 100 existing nuPlan logs and use the original route extracted from these logs as the `easy' route. `Hard' routes are generated from the same initial pose while maximizing the number of turns. For the \textbf{Lane \& Agent} task, as we no longer rely on the nuPlan maps, we dynamically adjust the difficulty of generated routes during DiT inpainting. We generate 100 scenarios. For each, we perform iterative inpainting starting from an initial 64m$\times$64m area, and perform a depth-first search on the updated lane graph at every inpainting step. `Hard' routes correspond to selecting the the next point to inpaint from to be the endpoint reachable with the most new turns, and `easy' routes are generated from endpoints with the fewest turns. We evaluate on route lengths of 100m (with a simulation time of 30 seconds) and 500m (150 seconds).  We consider both `easy' and `hard' traffic densities: `easy' is a single sample, and `hard' is the scene with the largest number of agents among 8 DiT samples.

\boldparagraph{Metrics} We evaluate the \textbf{Planner Failure Rate (PFR)} in SLEDGE using PDM-Closed~\cite{Dauner2023CORL}. This is the winner of the 2023 nuPlan challenge, and the state of the art for motion planning in the short, 15 second scenarios possible with existing data-driven simulators. The planner `fails' if it achieves less than 20\% of the route's total progress, goes in the wrong driving direction for more than 6m, goes off-road, or causes an at-fault collision. We also measure the \textbf{\#Turns} and \textbf{\#Agents}, which are proxies of the difficulty of the route and traffic.

\begin{table}[t!]
    \caption{\textbf{Simulation of PDM-Closed in SLEDGE.} Our simulator offers control over the route length, difficulty and traffic density. In several settings, we present new challenges for the existing state-of-the-art, leading to high failure rates of over 40\%.
    }
    \vspace{-0.0cm}
    \centering
    \resizebox{\linewidth}{!}{
    \setlength{\tabcolsep}{3pt}
    \begin{tabular}{c|c|ccc|ccc|ccc|ccc}
        \toprule
        \multicolumn{2}{c}{\textbf{Task}} $\rightarrow$ & \multicolumn{6}{|c}{\textbf{Lane $\rightarrow$ Agent}} & \multicolumn{6}{|c}{\textbf{Lane \& Agent}}\\
        \midrule
        \multicolumn{2}{c}{\textbf{Length}} $\rightarrow$ & \multicolumn{3}{|c}{\textbf{100 meters}} & \multicolumn{3}{|c}{\textbf{500 meters}} & \multicolumn{3}{|c}{\textbf{100 meters}} & \multicolumn{3}{|c}{\textbf{500 meters}} \\
        \midrule
        \textbf{Routes} & \textbf{Traffic} & {Turns} & {Agents} & {PFR} & {Turns} & {Agents} & {PFR} & {Turns} & {Agents} & {PFR} & {Turns} & {Agents} & {PFR} \\
        \midrule
        \multicolumn{2}{c|}{\textit{Replay}} & 0.89 & 57.40 & 0.06 & 3.29 & 102.34 & 0.26 & - & - & - & - & - & -\\
        \midrule
        \multirow{2}{*}{Easy} & Easy & \multirow{2}{*}{0.89} & 44.61 & 0.07 & \multirow{2}{*}{3.29} & 125.23 & 0.25 & 0.61 & 27.30 & 0.22 & 2.22 & 110.51 & 0.39 \\
        & Hard & & 56.44 & 0.11 & & 167.47 & 0.39 & 0.57 & 39.11 & 0.20 & 2.30 & 173.91 & 0.44 \\
        \midrule
        \multirow{2}{*}{Hard} & Easy & \multirow{2}{*}{1.18} & 44.66 & 0.14 & \multirow{2}{*}{4.20} & 128.79 & 0.28 & 1.23 & 27.14 & 0.29 & 3.66 & 107.11 & 0.45 \\
        & Hard & & 57.65 & 0.11 & & 170.87 & 0.44 & 1.07 & 39.03 & 0.30 & 3.82 & 169.66 & 0.49 \\
        \bottomrule
    \end{tabular}
    }
    \label{tab:sim}
    \vspace{-0.3cm}
\end{table}

\boldparagraph{Results} We show our results in \tabref{tab:sim}. Importantly, we note that setting up an evaluation with SLEDGE only requires a 3 GB download of our DiT-L checkpoint for the Lane \& Agent mode, and an additional 1 GB download of the nuPlan maps for the Lane $\rightarrow$ Agent mode, in contrast to the 2 TB of logs needed to set up nuPlan. When using easy routes and traffic, we observe similar PFRs for both replay-based and generative simulation despite this large compression factor. For replay, we observe over a 4$\times$ rise in PFR from 0.06 to 0.26 when extending the route from 100m to 500m. These are primarily due to PDM-Closed's inability to make lane changes or overtake slow vehicles, which are important planning behaviors that are not strongly penalized on current benchmarks like Val14~\cite{Dauner2023CORL}. In all settings, switching to hard routes increases the number of turns, which is more challenging than straight driving and in turn deteriorates the planning performance. In the most challenging settings with hard routes and traffic, the PFR increases to over 40\%. %
\section{Conclusion}
\label{sec:conclusion}

We present SLEDGE, a generative simulator for vehicle motion planning based on latent diffusion. We conduct several experiments to show that it is more realistic, compact, controllable, and diverse than other generative and replay-based approaches. Additionally, we establish several baselines and metrics for the generative simulation task. We hope our work can lay the foundation for accelerating progress in data-driven simulation and vehicle motion planning.

\boldparagraph{Limitations} Evaluating simulators is hard. We provide metrics for the lane graph generation sub-task, and preliminary experiments on testing rule-based planning, but using the simulator for other downstream tasks, such as reinforcement learning, will be important to showcase its full potential. For efficiency and robustness, we use a relatively small FOV and simulation radius, a simplistic lane representation consisting of only the centerline (assuming constant lane widths), and rule-based traffic behavior with IDM. These issues could be alleviated by further scaling of our model and extensions for learned motion behavior~\cite{Zhong2023CORL, Sun2023ARXIV}. However, like other diffusion models, the compute requirements of our approach are already high. We see value in improving the efficiency of SLEDGE through relevant techniques for accelerating diffusion models~\cite{Liu2022ARXIV,Luo2023ARXIV,Sauer2024ARXIV,Lin2024ARXIV}.

\clearpage

\boldparagraph{Acknowledgments} This work was supported by the ERC Starting Grant LEGO-3D (850533), the DFG EXC number 2064/1 - project number 390727645, the German Federal Ministry of Education and Research: Tübingen AI Center, FKZ: 01IS18039A and the German Federal Ministry for Economic Affairs and Climate Action within the project NXT GEN AI METHODS. We thank the International Max Planck Research School for Intelligent Systems (IMPRS-IS) for supporting Kashyap Chitta and Daniel Dauner. We also thank Agniv Sharma for providing his reimplementation of HDMapGen, Bernhard Jaeger for proofreading, and the nuPlan team for open-sourcing their dataset and simulation tools to the community.
\bibliographystyle{splncs04}
\bibliography{src/bibliography_long,src/bibliography,src/bibliography_custom}
\end{document}